\documentclass[10pt]{article}

\usepackage{amsmath}
\usepackage{amssymb}
\usepackage{wasysym}
\usepackage{graphicx, subfigure}
\usepackage{epstopdf}
\usepackage{cite}

\usepackage{color} 

\usepackage[labelfont=bf,labelsep=period,justification=raggedright]{caption}

\makeatletter
\renewcommand{\@biblabel}[1]{\quad#1.}
\makeatother

\usepackage{hyperref}

\makeatother

\begin{document}
\begin{flushleft}
\textbf{\Large{Capturing natural-colour 3D models of insects for species discovery
and diagnostics}}{\Large{ }}
\par\end{flushleft}{\Large \par}

\begin{flushleft}
Chuong V. Nguyen$^{1,\ast}$, David R. Lovell$^{1}$, 
Matt Adcock$^{1}$, John La Salle$^{2,3}$ \\
 \textbf{{1} CSIRO Computational Informatics, Canberra, ACT, Australia}\\
 \textbf{{2} CSIRO Ecosystem Sciences, Canberra, ACT, Australia}\\
 \textbf{{3} Atlas of Living Australia, Canberra, ACT, Australia}\\
 \textbf{$\ast$ E-mail: chuong.nguyen@csiro.au } 
\par\end{flushleft}

\section*{Abstract}

Collections of biological specimens are fundamental to scientific
understanding and characterization of natural diversity---past, present
and future. This paper presents a system for liberating useful information from
physical collections by bringing specimens into the digital domain
so they can be more readily shared, analyzed, annotated and compared.
It focuses on insects and is strongly motivated by the desire to accelerate
and augment current practices in insect taxonomy which predominantly 
use text, 2D diagrams and images to describe and characterize
species. While these traditional kinds of descriptions are informative and useful,
they cannot cover insect specimens ``from all angles'' and precious specimens
are still exchanged between researchers and collections for this reason.
Furthermore, insects can be complex in structure and pose many challenges 
to computer vision systems. 
We present a new prototype for a practical, cost-effective system
of off-the-shelf components to acquire natural-colour 3D models of insects
from around 3mm to 30mm in length. (``Natural-colour'' is used
to contrast with ``false-colour'', i.e., colour generated from,
or applied to, gray-scale data post-acquisition.) Colour images are captured
from different angles and focal depths using a digital single lens
reflex (DSLR) camera rig and two-axis turntable. These 2D images are
processed into 3D reconstructions using software based on a visual
hull algorithm. The resulting models are compact (around 10 megabytes), afford
excellent optical resolution, and can be readily embedded
into documents and web pages, as well as viewed on mobile devices. The system
is portable, safe, relatively affordable, and complements the sort
of volumetric data that can be acquired by computed tomography. This system provides
a new way to augment the description and documentation of insect species
holotypes, reducing the need to handle or ship specimens. It opens
up new opportunities to collect data for research, education, art,
 entertainment, biodiversity assessment and biosecurity control.

\section*{Introduction}

Technology has a critical role to play in accelerating the understanding
of biological diversity and, for decades, scientists have strived to create 
accurate 3D duplicates of plants and animal specimens \cite{Mcnally1993}. 
This paper describes a novel method of using technology to liberate
information about physical specimens by bringing them into the digital
domain as natural-colour 3D models---consistent with ideas and directions
articulated by several other authors \cite{Godfray2002,Salle2009,Beaman2012,Blagoderov2012,Mantle2012,Smith2012,Ang2013,Balke2013,Johnson2013}.
In particular, the proof of concept system we present fits well with
the suggestion of Wheeler \emph{et al.} \cite{Wheeler2012} to ``engineer
and deploy a network of automated instruments capable of rapidly creating
3D images of type specimens'' as part of a larger strategy of dealing
with the massive backlog of insect types that are not yet digitized
in any form. 
High resolution 3D scans, as well as being useful as versatile replicas, 
also have the potential to act as a common frame of reference for 
other data relating to the original insect such as annotations, auxiliary image collections, 
and measurements. These additional aspects are vital for the ways taxonomists convey 
the various morphological characters that distinguish a new species from those previously discovered.

Our work is focused on the digitization of insect species, building
on research and development at the Australian National Insect Collection
(ANIC) which currently holds over 12 million specimens, and is growing
by around 100,000 specimens every year. Our mission is to enable high-quality
3D models of insects to be acquired quickly and cheaply, for
ANIC to use as a component of its digitization strategy. 
Like many Natural History collections around the globe, the ANIC maintains 
many (thousands) Holotypes - each the single specimen of a species that is 
used to define the characteristic features of that species. Holotypes exist 
as a physical object carefully protected from damage through handling. 
Digital colour 3D models of sufficient detail will enable collections managers 
to liberate these precious specimens for the research work they are intended to fulfill. 

Micro Computed
Tomography (Micro CT) is currently a key method \cite{Metscher2009, Faulwetter2013}, 
able to create micron-accurate
volumetric models of millimeter-scale objects and their internal structure.
However, like recent 3D reconstructions from scanning electron microscope
(SEM) micrographs \cite{Akkari2013,Koon-BongCheung2013}, Micro CT
is unable to capture important information about the surface of the
object: its natural colour. Exposure and reconstruction times can be long
(tens of hours) and, as an X-ray imaging method, Micro CT generally
demands special safety equipment. Current systems cost in the hundred-thousand
dollar range and, while more compact desktop models are available,
these are still not especially portable.

The inability of X-ray based methods for insect digitization
to capture colour led us to consider image-based
3D reconstruction techniques as reviewed in \cite{Hartley2004,Szeliski2010}.
These methods have been successfully applied to the reconstruction
of 3D cityscapes and other (generally fairly simple) objects \cite{Furukawa2010,Hernandez2010,Tola2011}.
Some small biological specimens have been digitized \cite{Atsushi2011,Gallo2013,Chalmers2012}
but the methods used do not specifically cater for the complex structures
and challenging surface optical properties of insects. Human-in-the-loop
approaches have been proposed for insect modeling \cite{Murakawa2006}
as have methods (limited to simple insect geometries) for inferring
3D insect shape from a single 2D image \cite{Zhang2010}.
Experiments \cite{Felicisimo2012, Kuzminsky2012} with laser scanning systems like \cite{NextEngine2013}
have suggested that this approach has difficulties with the fine structures
and the small scale of many insects, as well as reflective, transparent
or iridescent surfaces.

One way to avoid these difficulties is to
steer clear of 3D reconstruction altogether and simply present 2D
images obtained from different viewing angles \cite{Ortery2013}.
While this method of 3D visualization is popular for museum collections
it does not provide the quantitative information (e.g., 3D morphology)
needed to analyze and compare insect specimens. Furthermore large
amounts of data are involved: many high-resolution images are needed
to give a convincing illusion of looking at an actual 3D object. This
makes smooth, realistic interaction difficult and precludes straightforward
email exchange or embedding of the object data.

In summary, there is a lack of existing systems that could
capture the 3D structure and surface optical properties of small,
intricate insect specimens at sufficient resolution for ANIC and other
collections to digitize, share, analyze and compare their holdings.
The rest of the paper describes our prototype system and its operation, 
and how it has achieved these design objectives.

\section*{Materials and methods}

\begin{figure}[!ht]
\begin{center}
\href{http://www.youtube.com/watch?v=THvfu6shJjg&list=UUak3NZxjNnWip327vYL8RLA}
{\includegraphics[width=12.35cm]{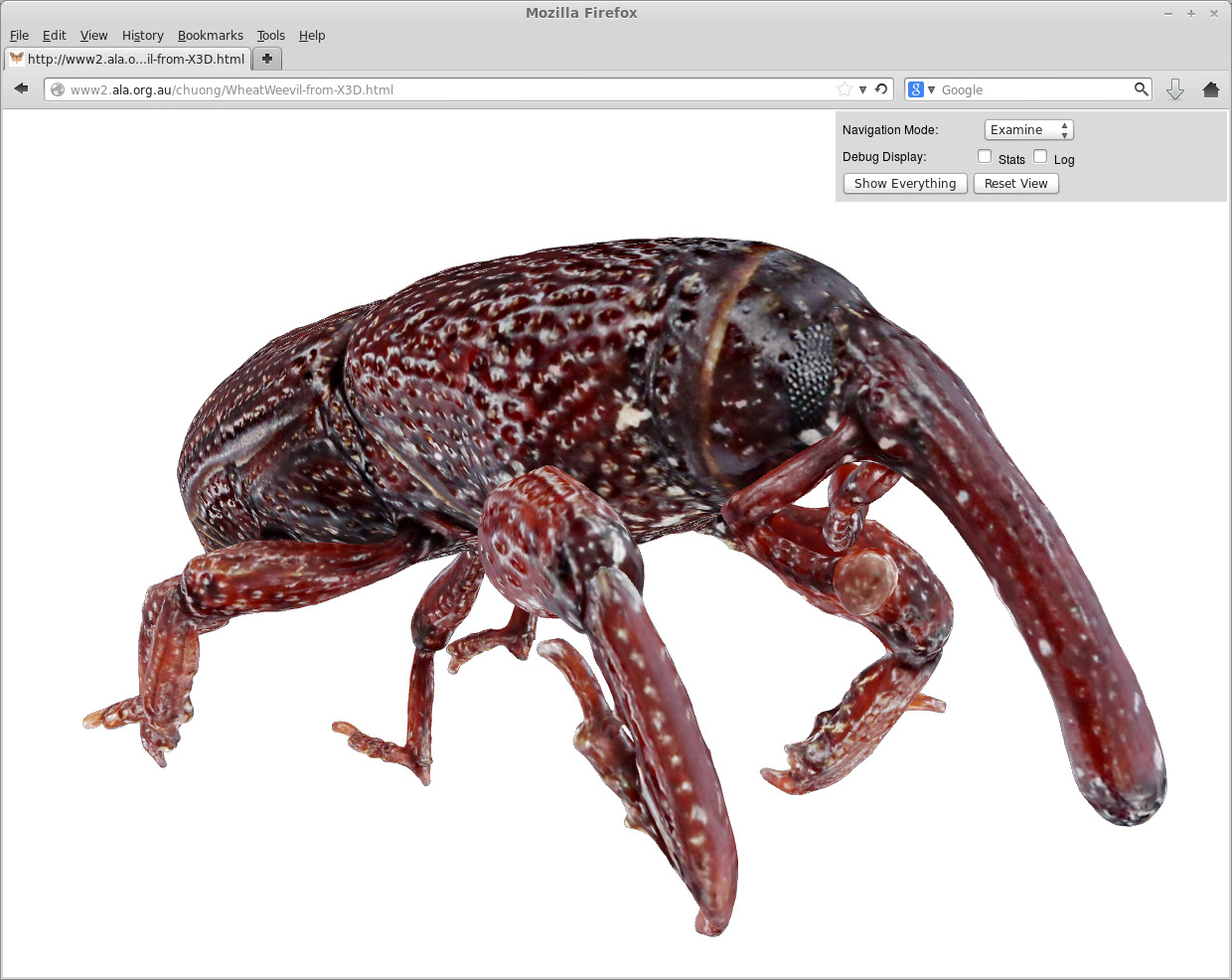} }
\end{center}
\caption{\label{fig:process_video} \textbf{3D visualisation of a granary weevil on web as part of a 
video showing an overview of the 3D scanning process.} 
Click on the figure, or go the link at \cite{Nguyen2014i} to view the video.}
\end{figure}

Here we provide overviews of the digitization process and equipment. 
A video \cite{Nguyen2014i} as depicted in Figure \ref{fig:process_video} shows 
the main components of the system and the digitization process in action.

\subsection*{Process overview}

\begin{figure}[!ht]
\begin{center}
\includegraphics[width=12.35cm]{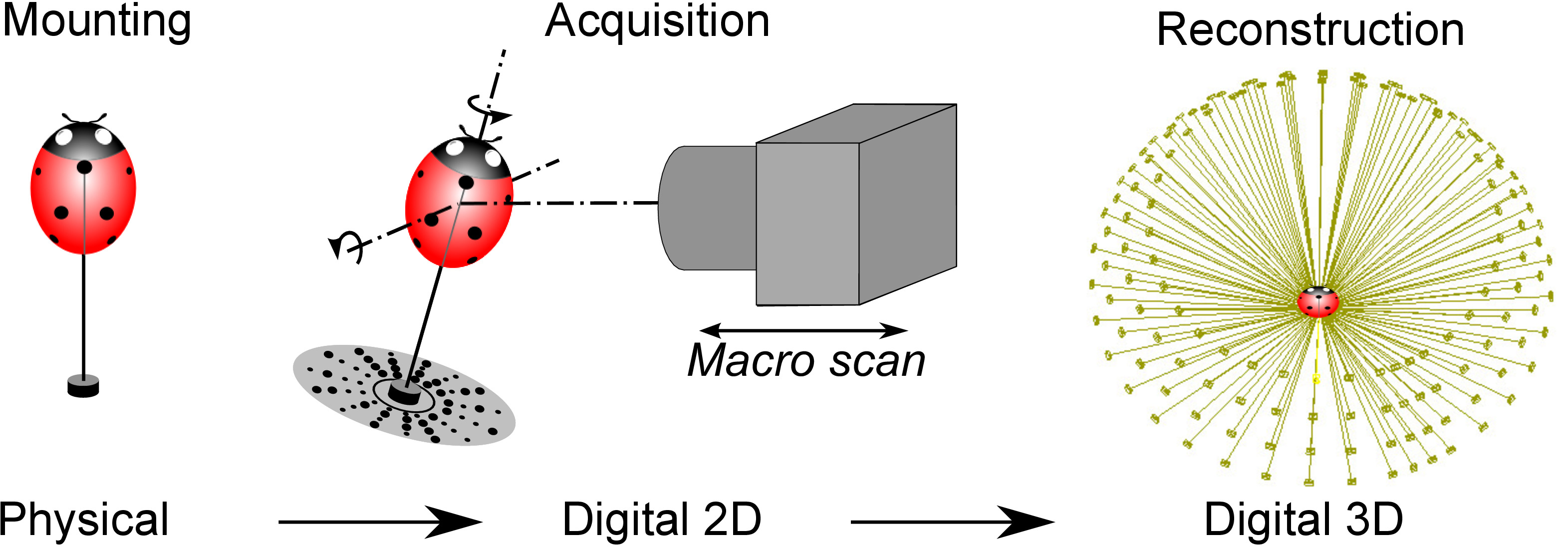} 
\end{center}
\caption{\label{fig:process} \textbf{The three main steps to create a natural-colour
3D model of specimen.} The steps are mounting the insect onto a pin,
acquisition of 2D images of the specimen at different poses, then reconstruction of 
a single 3D model from those multiple images.}
\end{figure}

In high-level terms, our system and work-flow involve three main steps
(Figure~\ref{fig:process}):
\begin{description}
\item [{Mounting:}] the physical specimen is pinned onto a pre-printed
mat used later by the reconstruction software to estimate camera pose (viewing
angle and position).
\item [{Acquisition:}] 2D images of the specimen are automatically acquired
from different orientations (and focal depths for small insects).
This step marks the transition from the physical to the digital domain.
\item [{Reconstruction:}] in which a 3D model is inferred from multiple
2D images. For small insects, this involves multi-focus image stacking
before the general steps of extracting camera pose, shape and colour.
\end{description}
The system has two modes of acquisition, depending on the specimen
size.  Insects larger than 10mm are captured in \emph{normal-mode} in which the depth
of focus of the normal DSLR camera lens is enough to keep the whole
specimen in focus at any viewing angle. Insects smaller than 10mm are 
captured in \emph{macro-mode}
using a high-magnification lens. Because of the shallow depth of focus
of this lens, multiple images are captured at different distances
from the specimen and processed into a single in-focus image.

\subsection*{Equipment overview}

\begin{figure}[!ht]
\begin{center}
\textbf{\includegraphics[width=12.35cm]{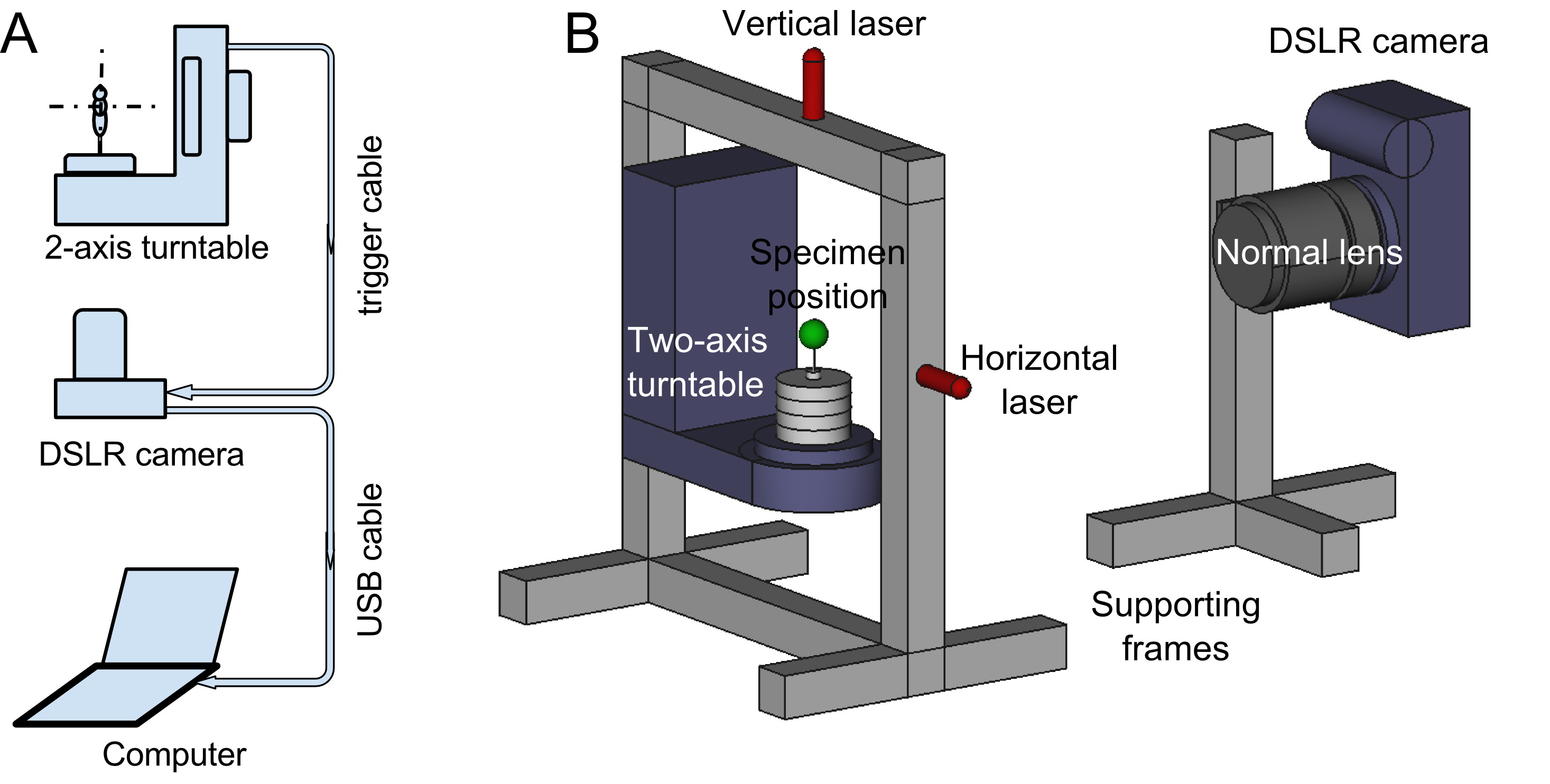}} 
\end{center}
\caption{\label{fig:normal}\textbf{ Connections (A) and hardware (B) for \emph{normal-mode}
image acquisition.} The green sphere marks the center of rotation and
mounting location of specimens. The turntable is the master device
that triggers the camera after rotating to predetermined pan and tilt
angles. Images can be stored in camera memory or transferred directly
to the computer as they are acquired. }
\end{figure}

\begin{figure}[!ht]
\begin{center}
\textbf{\includegraphics[width=12.35cm]{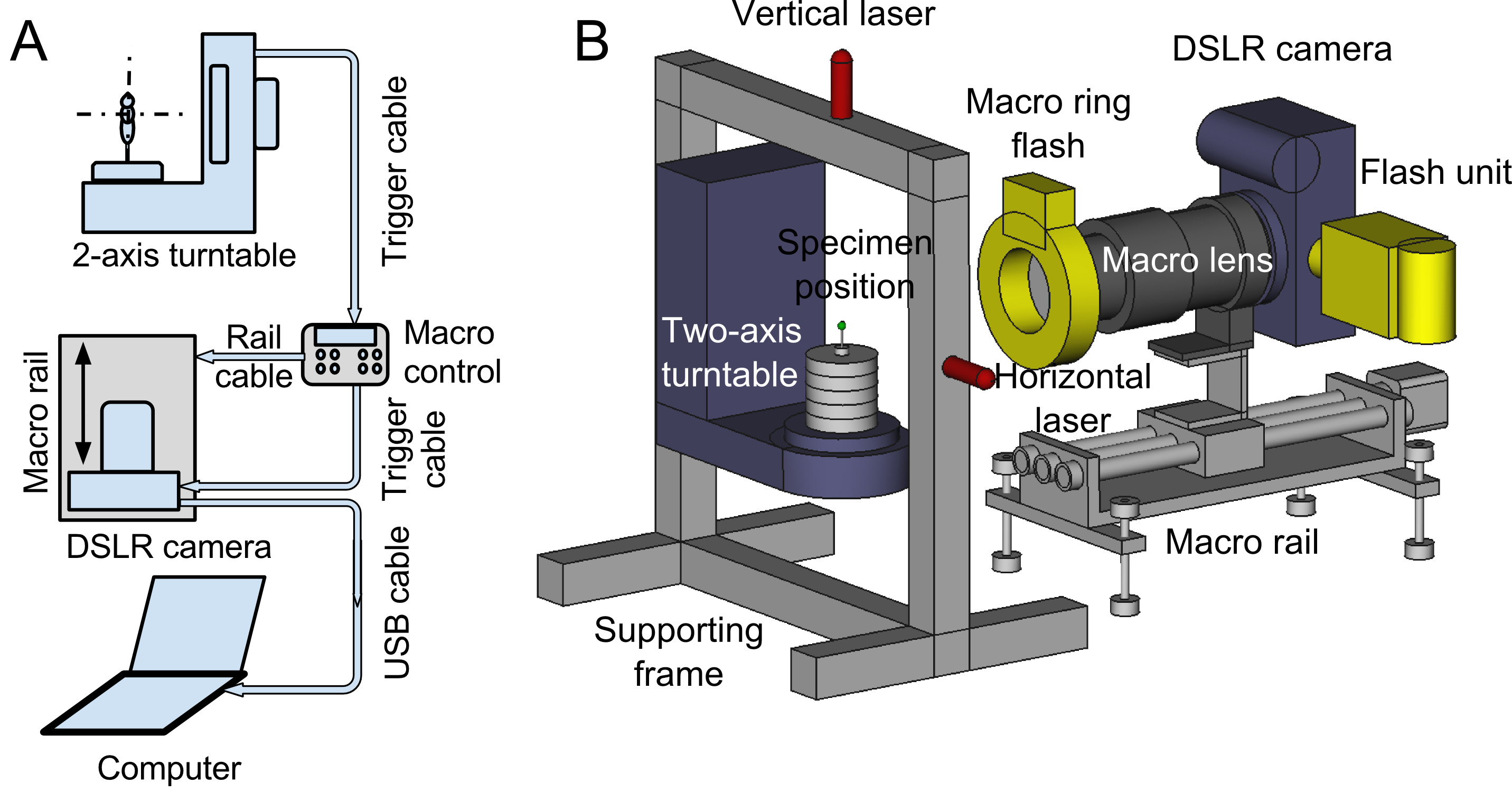}}
\end{center}
\caption{\label{fig:macro} \textbf{Connections (A) and hardware (B)
for \emph{macro-mode} image acquisition.} The macro lens, macro ring
flash and macro-rail are needed for capturing high-magnification and
depth-extended images of small insects. At each rotation step, the
turntable triggers the control box of macro-rail. The macro-rail then
moves to a set of predetermined positions. At each position, the control
box triggers the camera to capture an image. }
\end{figure}

Figures~\ref{fig:normal} and~\ref{fig:macro} show normal- and
macro-mode setups. The main hardware components of the system are:
\begin{itemize}
\item A two-axis turntable to present views of the specimen from different
angles of rotation
\item A macro-rail to vary the distance between the camera and specimen
in macro-mode
\item A camera and flash. 
\item Two laser pointers for specimen alignment
\item A computer for 2D image processing and 3D reconstruction.
\end{itemize}

It is noted that in macro-mode our system uses a macro-rail to capture multi-focus 
images exactly at predefined depths, as opposed to refocusing the camera lens. 
A camera flash is needed to eliminate motion blur due to camera shutter's vibration
when capturing at high magnification.

To minimize cost and development time we sought to use off-the-shelf
components wherever practicable. These are described in detail in
the Supplementary Information.

\subsection*{Step 1: Mounting}

Collections usually store and display insects larger than $\sim10$mm
by pinning them so that the insect's long axis is horizontal and the
pin vertical. Insects smaller than $\sim10$mm are usually either pinned or glued in cards. 
This paper however focuses on pinned insects and issues arising from this mounting method.
Pinning insects horizontally allows many
insects to be stored in wide, flat display drawers but creates a few
problems for our system:
\begin{itemize}
\item The pin becomes part of the 3D model and must be edited or segmented out in post-reconstruction
\item Editing can often not fully remove evidence of the pin
\item Images of the underside of the specimen can be difficult or impossible
to capture, leading to an incomplete 3D model.
\end{itemize}

Re-pinning the insect so its long axis is vertical helps with image
acquisition but risks damaging the specimen, including parts, 
such as genitalia,  that are important for the identification of some species.
For some specimens, 
these affected parts can be isolated through dissection and scanned separately.

\begin{figure}[!ht]
\begin{center}
\textbf{\includegraphics[width=17.35cm]{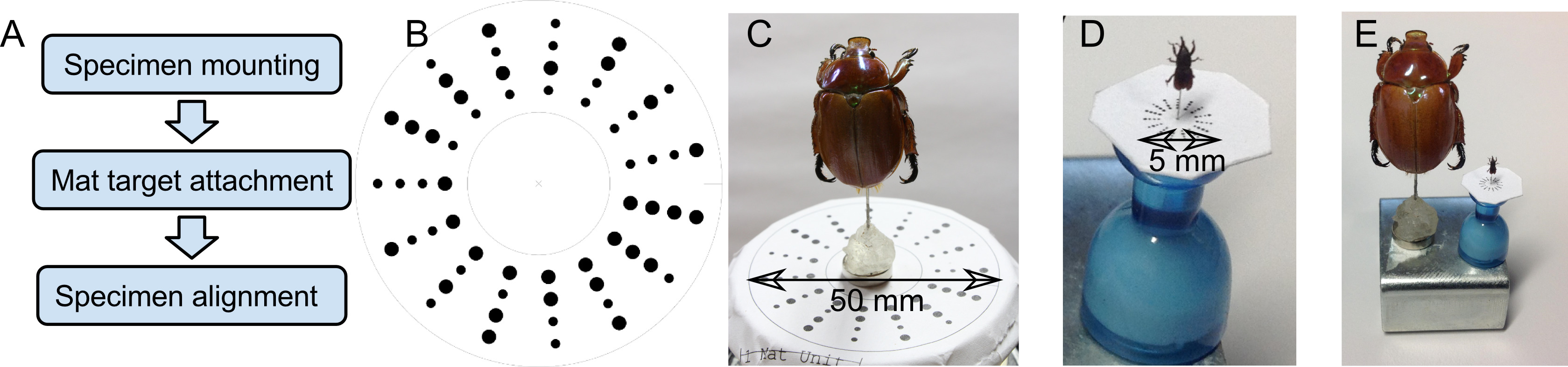}} 
\end{center}
\caption{\label{fig:mounting} \textbf{Preparing insect specimen for scanning}. A) Steps
to prepare insect specimens for image capturing. B) A special mat
target needs to be attached to a scanned specimen for 3DSOM$^{TM}$
software to estimate of camera viewing position and angle. C) For
a large insect such as this 30mm long Christmas beetle, the pin is
glued to a $\diameter$10mm rare-earth disk magnet which is in turn attached
to a $\diameter$50mm mat target. D) For a small insect such as this 3mm
long granary weevil, the micro pin is glued to a $\diameter$5mm mat
target. E) shows comparison in size of the two specimens. }
\end{figure}

After the specimen is pinned, the pin is glued to a small magnet (Figure
\ref{fig:mounting}C) that will hold the pin in position on the
turntable. Next, a specially patterned mat (Figure \ref{fig:mounting}B), 
required by the reconstruction software (3DSOM\texttrademark{}  
\cite{AdamBaumbergCreativeDimensionSoftwareLtd}),
is attached to provide information about camera pose and position 
relative to the specimen. Generally the suitable size of the 
pattern is about one to two times the length of the insect to be scanned. 
Scanning smaller insects requires smaller patterns to be printed.
Currently, modern laser printers with 1200 dpi printing resolution 
can produce patterned mats as small as 5mm in diameter. 
Printing smaller patterns that are sharp enough to be recognised 
by the reconstruction software is currently a technical challenge.

Finally, the whole assembly is placed on the two-axis turntable and 
positioned (with the assistance of horizontal and vertical laser pointers) 
so the specimen is centered on the intersection of the axes of tilt and rotation. 
The lasers are aligned to the rotation axes of the turntable. 
A specimen is manually aligned to each of the laser beams such that 
each beam hits the centre of the insect's body.

\subsection*{Step 2: Acquisition}

This is the point at which the physical specimen enters the digital
domain.

\begin{figure}[!ht]
\begin{center}
\textbf{\includegraphics[width=17.35cm]{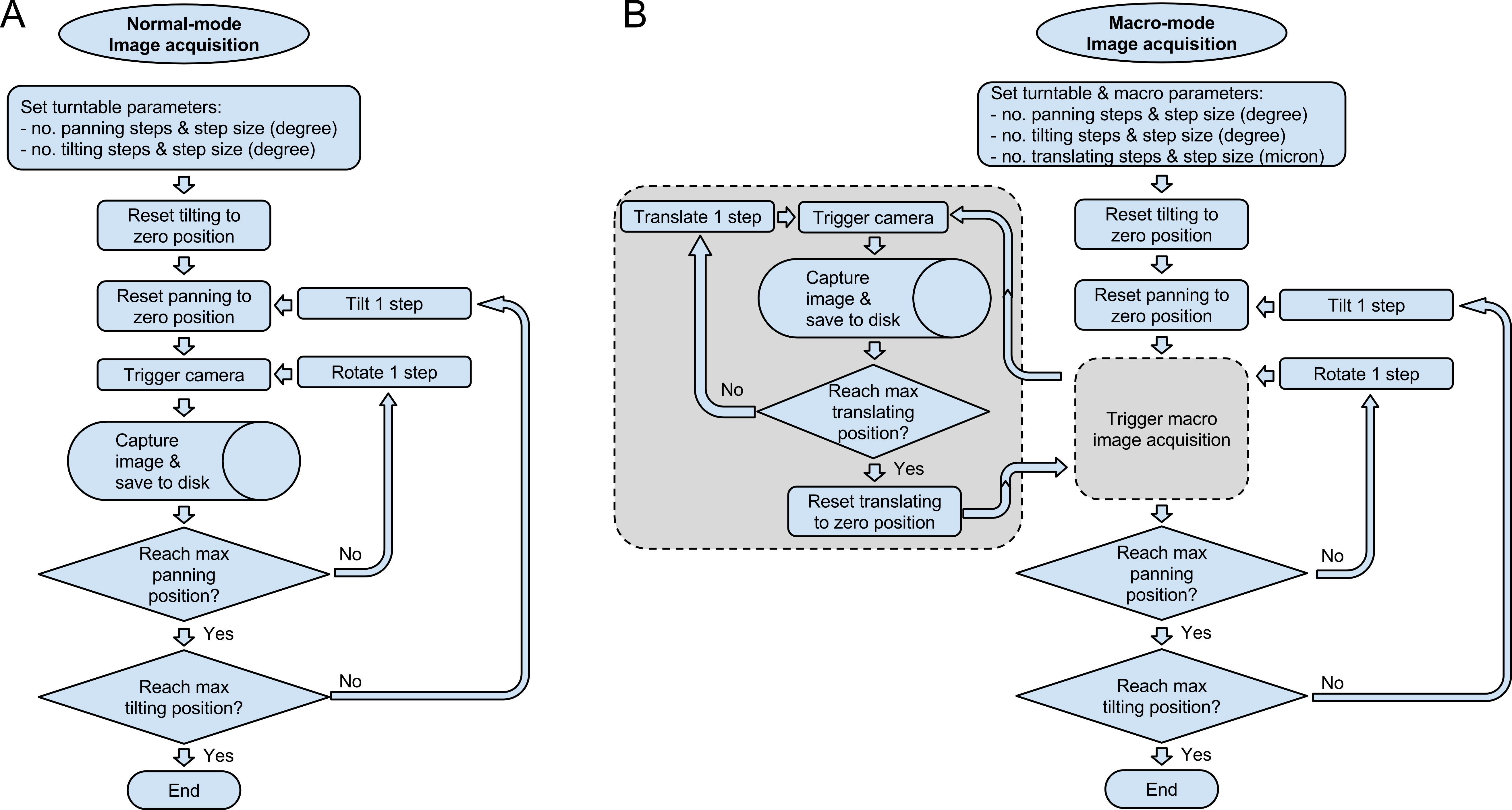}} 
\end{center}
\caption{\label{fig:flowcharts} \textbf{Automated image acquisition process.}
A) Normal-mode. B) Macro-mode. }
\end{figure}

In essence, the acquisition process is about automatically obtaining
2D images of the specimen in different poses. As far as the relationship
between the camera and specimen goes, this system has three degrees
of freedom: pan, tilt and (in macro-mode) distance along the specimen-camera
axis. With the specimen mounted at the intersection of the pan and
tilt axes of the turntable, this amounts to rotating the turntable
through a range of pan and tilt angles, capturing an image at each
step (Figure~\ref{fig:flowcharts}A). In macro-mode there is
an additional ``inner loop'' of translating the camera to acquire
partially focused images at different distances from the specimen
for later processing into a single image with all parts of the specimen
fully in focus (Figure~\ref{fig:flowcharts}B).

There are many ways to automate the acquisition process. The desire
to use off-the-shelf components led us to use the GigaPan\texttrademark{}
Panorama Robot EPIC 100 \cite{GigaPanSystems2013} for mounting the
\emph{specimen}. The GigaPan\texttrademark{} is designed for mounting
and controlling a \emph{camera}---and this led to the GigaPan\texttrademark{}
robot also acting as the acquisition controller. In other words, it is the turntable
that triggers the macro-rail. The macro-rail moves and triggers the camera which triggers
its flash and takes an image. The Supplementary Information contains more detail about
this set-up.

In normal-mode, using rotation and axis tilt, the set-up captures 144 individual images.  
In macro-mode, the additional up to 31 images required at each step mean that the system 
can capture up to 4,464 separate images per specimen. Capturing more images is also possible.

\subsection*{Step 3: Reconstruction}

\begin{figure}[!ht]
\begin{center}
\textbf{\includegraphics[width=17.35cm]{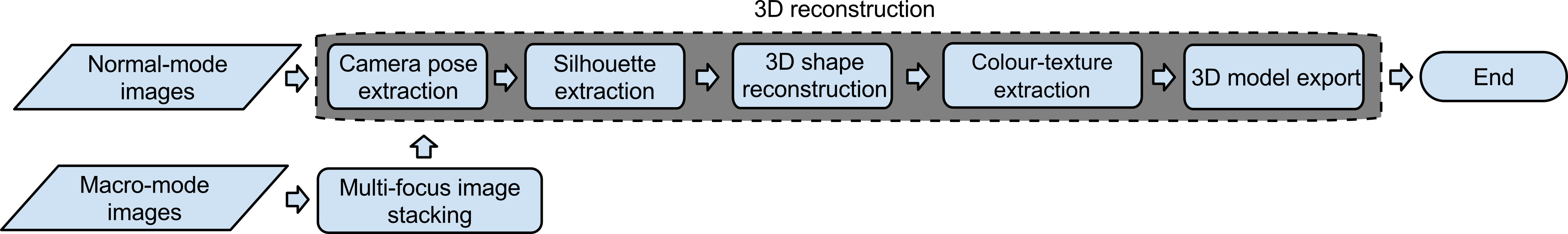}} 
\end{center}
\caption{\label{fig:image_pipeline} \textbf{Image processing pipeline for normal-mode
and macro-mode images.} Macro-mode images require an extra step to
stack each set of multi-focus images captured from the same viewing
angle (but at different depth distances) into a single in-focus image. }
\end{figure}

The third and final step of the digitization process is where 
the 2D digital information acquired from a physical specimen is manipulated to produce 
a 3D digital model (Figure~\ref{fig:image_pipeline}) .

In macro-mode, the stack of partially focused images acquired at different
specimen-camera distances must be combined into a single in-focus
image for a given viewing angle. We used Helicon Focus \cite{HeliconSoft2013}
for this because of its ability to exploit multiple CPU cores. Single
core open-source alternatives are available \cite{Hadley2010,D'Angelo2013}.

Armed with a set of in-focus 2D images of an object from different
viewing angles, there are two main 3D reconstruction techniques that
could be applied:

\begin{description}
\item [{Visual~hull}] (also known as \emph{volume carving}) algorithms
\cite{Laurentini1994,Franco2003} project the silhouette of the
object into a virtual volume at each viewing angle, carving away the
volume outside the silhouette to leave a 3D visual hull which approximates
the shape of the actual object. This approach does not recover concave surfaces,
but photo-consistency can be used to correct this to an extent\cite{Kutulakos2000}.
The extent of improvement by photo-consistency is limited for some insects 
due to strong speculiar reflections 
on the outer-surface and fine body structures such as legs, antennae, spikes and hairs.

\item [{Multi-view~stereo}] algorithms generally rely on photo-consistency
measures to identify the location of common features seen in different
views \cite{Goesele2006, Seitz2006} and can also incorporate silhouette
information \cite{Sinha2005}.
\end{description}

Both strategies are computationally intensive and the computational demands
increase with reconstruction resolution. Image clustering \cite{Furukawa2010a,Furukawa2010}
and improved feature descriptors \cite{Tola2011} have been previously proposed
to enable reconstructions to better exploit the very high image resolution
produced by professional photography cameras.

Our initial investigations indicated that the visual-hull-based method could 
more accurately reconstruct some of the thin structures found in insects (e.g., legs, antennae,
wings) and insect surfaces with strong specular reflections. 3DSOM\texttrademark{}  
\cite{AdamBaumbergCreativeDimensionSoftwareLtd} was used to provide off-the-shelf
visual-hull-based reconstruction as it produced the best quality output
of the different approaches \cite{Furukawa2010a,Snavely2010,Forbes2009}.

Figure~\ref{fig:image_pipeline} sets out the detail of the reconstruction
process, including the extraction of the camera pose in each input
image. 3DSOM\texttrademark{} initially estimates this information
from the target pattern captured in the image and further refines these
estimates during 3D reconstruction. Specimen silhouettes are extracted
from input images. Once the 3D geometry of the specimen's surface
is reconstructed, texture colour is extracted from the images and
added to the model. The resulting 3D model can be then exported to
different formats---including HTML (with WebGL, Flash or Java), X3D,
3DS (AutoDesk), and STL (STereoLithography)---for subsequent viewing,
analysis or embedding into documents. X3D is a convenient format 
as it is supported by popular 3D visualisation software, and a X3D file 
can included as an embedded object or as XML inline in an HTML5 file 
for 3D web visualisation. InstantReality's \cite{instantreality2014} 
tool ``aopt" can perform this conversion X3D to 3D-supported HTML automatically.

\section*{Results and discussion}

\begin{figure}[!ht]
\begin{center}
 \subfigure{%
 \href{http://www2.ala.org.au/chuong/WheatWeevil-from-X3D.html} 
 {\includegraphics[width=3.2cm]{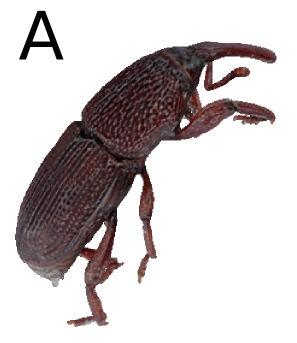}}
 }%
 \subfigure{%
 \href{http://www2.ala.org.au/chuong/Bee-from-X3D.html} 
 {\includegraphics[width=3.7cm]{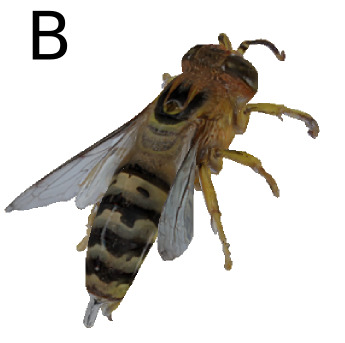}}
 }%
 \subfigure{%
 \href{http://www2.ala.org.au/chuong/LongHornBeetle-from-X3D.html} 
 {\includegraphics[width=3.35cm]{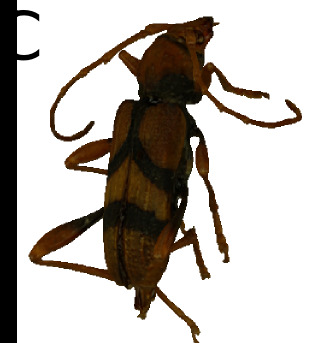}}
 }%
 \subfigure{%
 \href{http://www2.ala.org.au/chuong/Christmas_Beetle-from-X3D.html} 
 {\includegraphics[width=3.7cm]{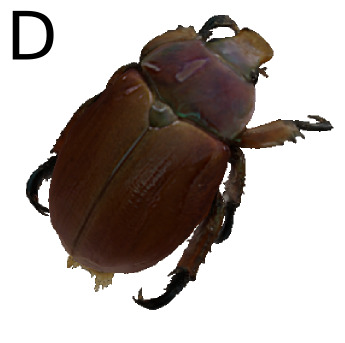}}
 }%
 \subfigure{%
 \href{http://www2.ala.org.au/chuong/Weevil-from-X3D.html} 
 {\includegraphics[width=2.9cm]{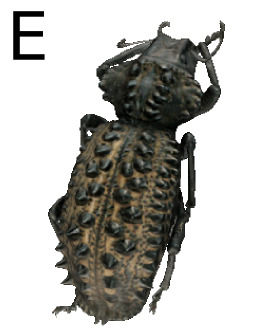}}
 }\\%
 \subfigure{%
 \href{http://www2.ala.org.au/chuong/} 
 {\includegraphics[width=17.35cm]{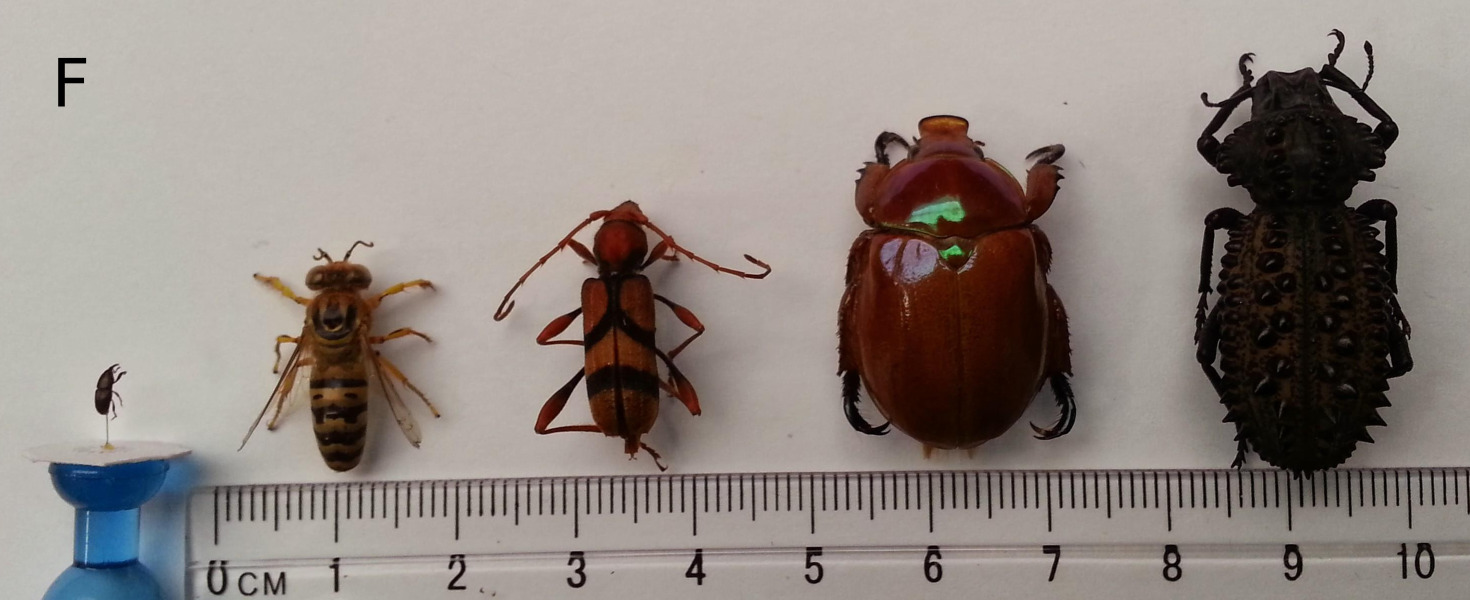}}
 }%
\end{center}
\caption{\label{fig:bugs} \textbf{Various 3D insect models.} 
Click on individual figure or go to the link at \cite{Nguyen2013} 
to interact with the 3D models or 
\cite{Nguyen2014a, Nguyen2014b, Nguyen2014c, Nguyen2014d, Nguyen2014e, Nguyen2014f, Nguyen2014g, Nguyen2014h} to download. 
Top: 3D models of the insects with
natural-colour texture, scaled to have similar sizes. They are 
A) a granary weevil (\emph{Sitophilus granarius}), 
B) a sand wasp (\emph{Bembix sp.}), 
C) a longhorn beetle (\emph{Aridaeus thoracicus}), 
D) a Christmas beetle (\emph{Anoplognathus viriditarsis})
and E) a amycterine ground weevil (\emph{Gagatophorus draco}). 
Bottom: F) A photograph of the real insect specimens of the 3D models captured. }
\end{figure}

Figure~\ref{fig:bugs} shows high-resolution natural-colour 3D models of
insects ranging from 3mm to 30mm in length.
These 3D insect models are also available for interactive viewing at \cite{Nguyen2013}
and can be downloaded at \cite{Nguyen2014a, Nguyen2014b, Nguyen2014c, Nguyen2014d, Nguyen2014e, Nguyen2014f, Nguyen2014g, Nguyen2014h}.
The smallest of these---the 3mm granary weevil---proved challenging
to resolve due to an out-of-focus problem when its images were captured at $2\times$ magnification.
The 3D model of granary weevil was obtained from images captured in macro-mode, while 3D models of 
larger insects were obtained from images captured in normal-mode.
The 3D visualisation of insect models is based on the open-source X3DOM framework \cite{X3DOM2014}
which uses WebGL for plug-in-less display within a web browser 
(such as Firefox and Chrome). The file size of models, including 3D mesh and texture, 
depends on the desired visualisation quality and the complexity of the geometry and 
colour of the actual specimen. For the 3D models shown at \cite{Nguyen2013}, the file size 
ranges from 5 to 24 megabytes, with number of vertices from 80,000 to 130,000 and texture resolution from 4 to 16 megapixels.

\begin{figure}[!ht]
\begin{center}
\textbf{\includegraphics[width=12.35cm]{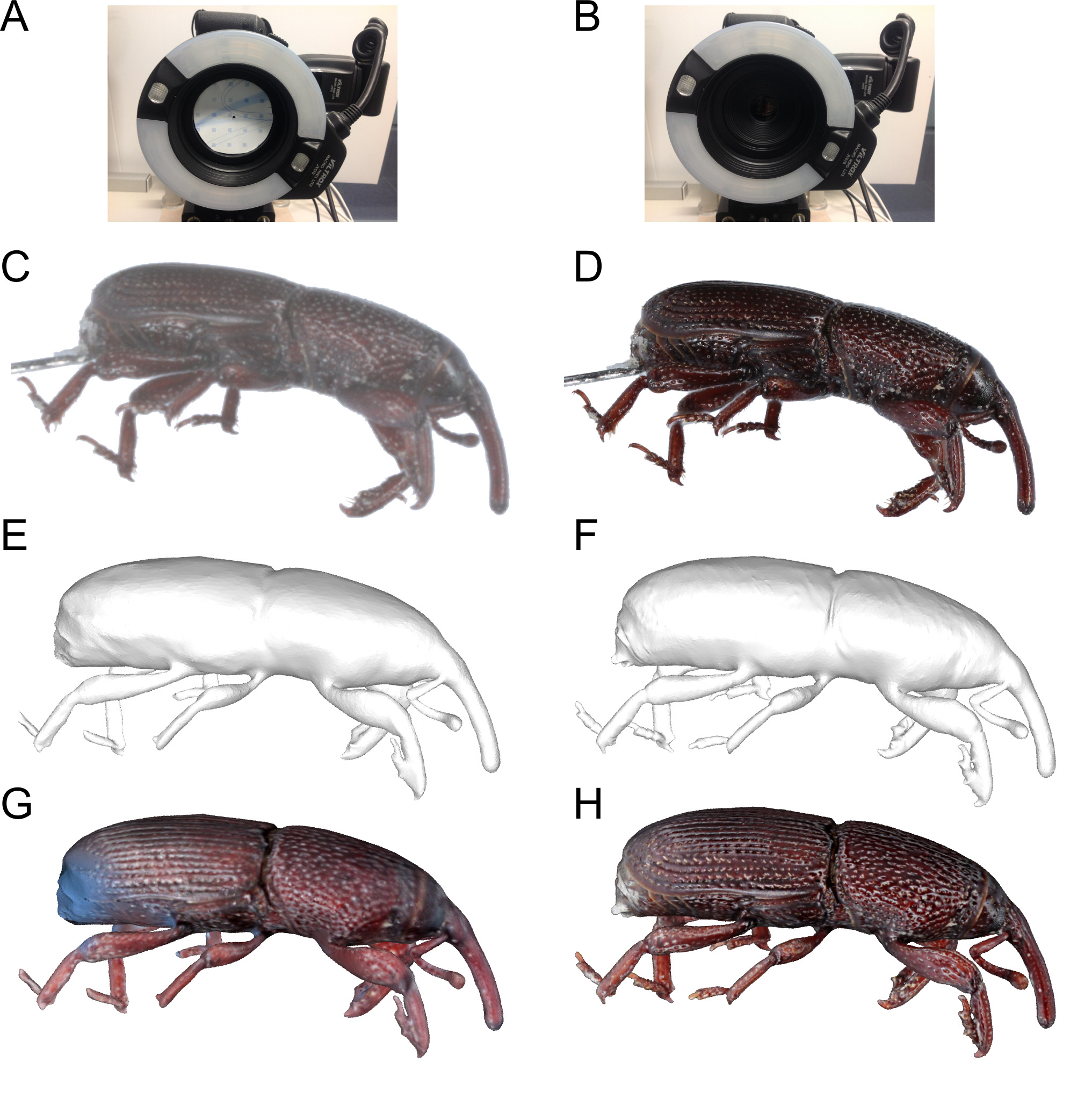}} 
\end{center}
\caption{\label{fig:aperture_vs_multifocus} \textbf{Comparison of natural-colour 3D reconstructions
 using (A) a small aperture and (B) a F/8 aperture with multi-focus image stacking}. 
A) shows an extra mask with a $\diameter$2mm hole
put in front of the lens to extend depth of focus as compared to B) an F/8
lens aperture. C) the resulting images
captured at the same angle by small aperture. D) multi-focus
image stacking from 31 partial-focus images captured at distances
0.25mm apart. E)-H) show screen shots of resulting 3D models
without and with texture colour. }
\end{figure}

Figure~\ref{fig:aperture_vs_multifocus} illustrates the effectiveness of macro-mode image acquisition 
as compared to normal-mode  image acquisition when applied to very small insects such as the granary-weevil.
A Canon EF-65mm macro lens was employed in both cases.
In normal-mode, a stencil with a $\diameter$2mm hole had to be attached immediately
in front of the camera (Figure~\ref{fig:aperture_vs_multifocus}A) 
to reduce the effective aperture and increase the depth of focus. 
In both cases a flash was used to mitigate the effects of wobble due to 
the camera shutter movement. With a flash, the exposure time of an image is 
effectively the very short duration of the flash when it triggers, 
and therefore it minimizes any motion blur.
Flash energy in macro mode was $\frac{1}{128}$ of full power and in normal-mode 
(for the $\diameter$2mm aperture) it was $\frac{1}{16}$ of full power. 
The results shown in Figure~\ref{fig:aperture_vs_multifocus}
clearly illustrate the improvements of macro-mode. The macro-mode model was reconstructed with 
 multi-focus stacking of 31 images from each view, each captured with an F/8 lens aperture
at increments of 0.25mm along the specimen-camera axis.

\begin{figure}[!ht]
\begin{center}
\textbf{\includegraphics[width=17.35cm]{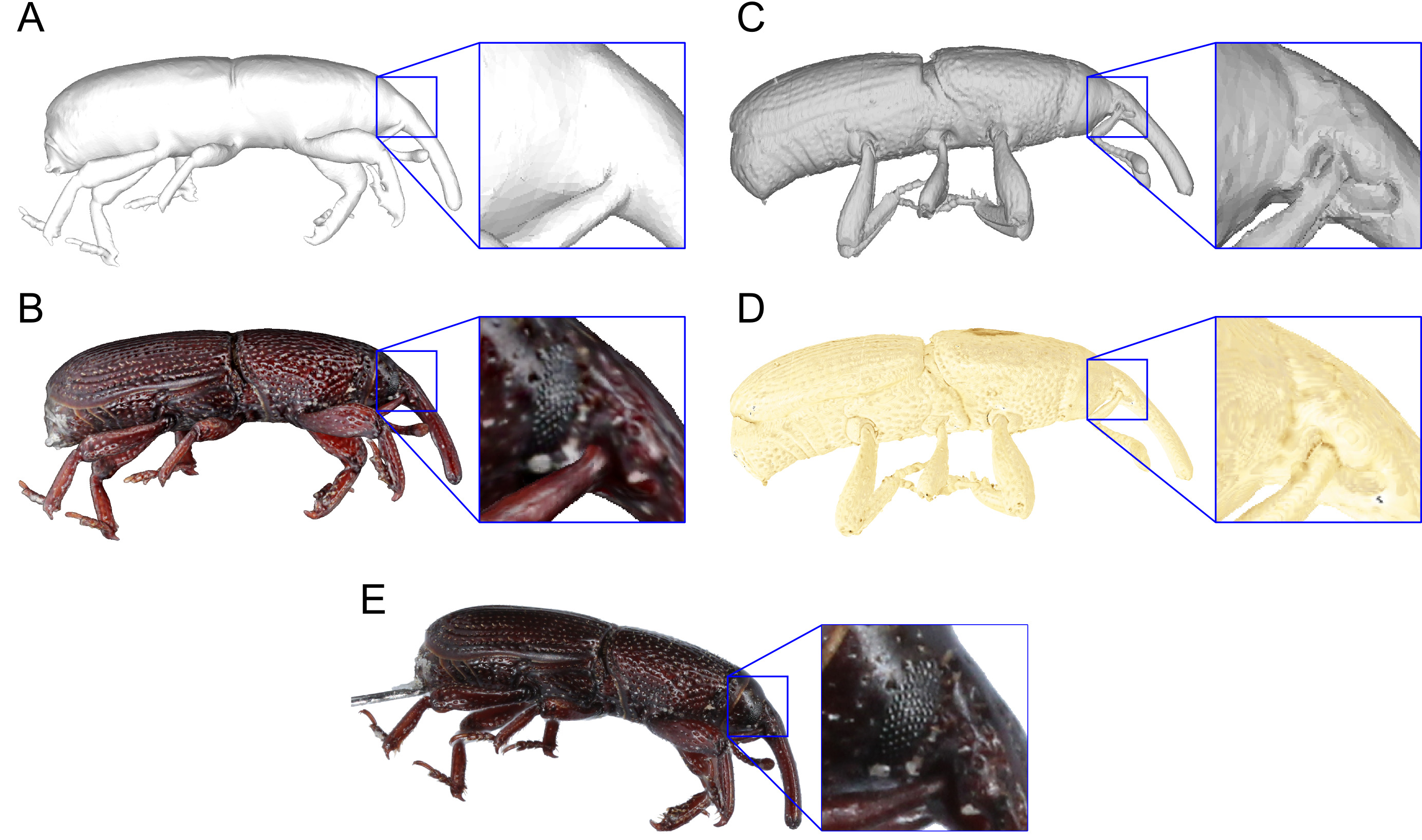}} 
\end{center}
\caption{\label{fig:colour_vs_Micro CT} \textbf{Comparison of a natural-colour 3D model, 
a Micro CT reconstruction and 2D image at a similar angle.} The surface geometry
of the natural-colour 3D model (A) is less detailed than the Micro CT model (C) 
and missed concavities such as the antenna socket shown in the enlarged inset of C.
However, the natural-colour 3D model can capture useful surface information such as 
the compound eye in the enlarged insect of B. False-colour Micro CT model (D) 
and a 2D image (E) are shown for comparison.}
\end{figure}

Figure~\ref{fig:colour_vs_Micro CT} provides a qualitative comparison
of a natural-colour 3D model obtained using our system and a Micro CT
model of a different specimen of the same species.
While the 5.7$\mu$m resolution Micro CT clearly captures more details
of the surface geometry than our optical approach (including the missing
antenna socket in inset A), there are features that
 it cannot resolve at these resolutions because they are to do with
variation in the colour of the specimen (e.g., the compound eye in
inset B). One option could be to develop ways to combine the strengths
of both approaches: fertile ground for further research.

\begin{figure}[!ht]
\begin{center}
\textbf{\includegraphics[width=12.35cm]{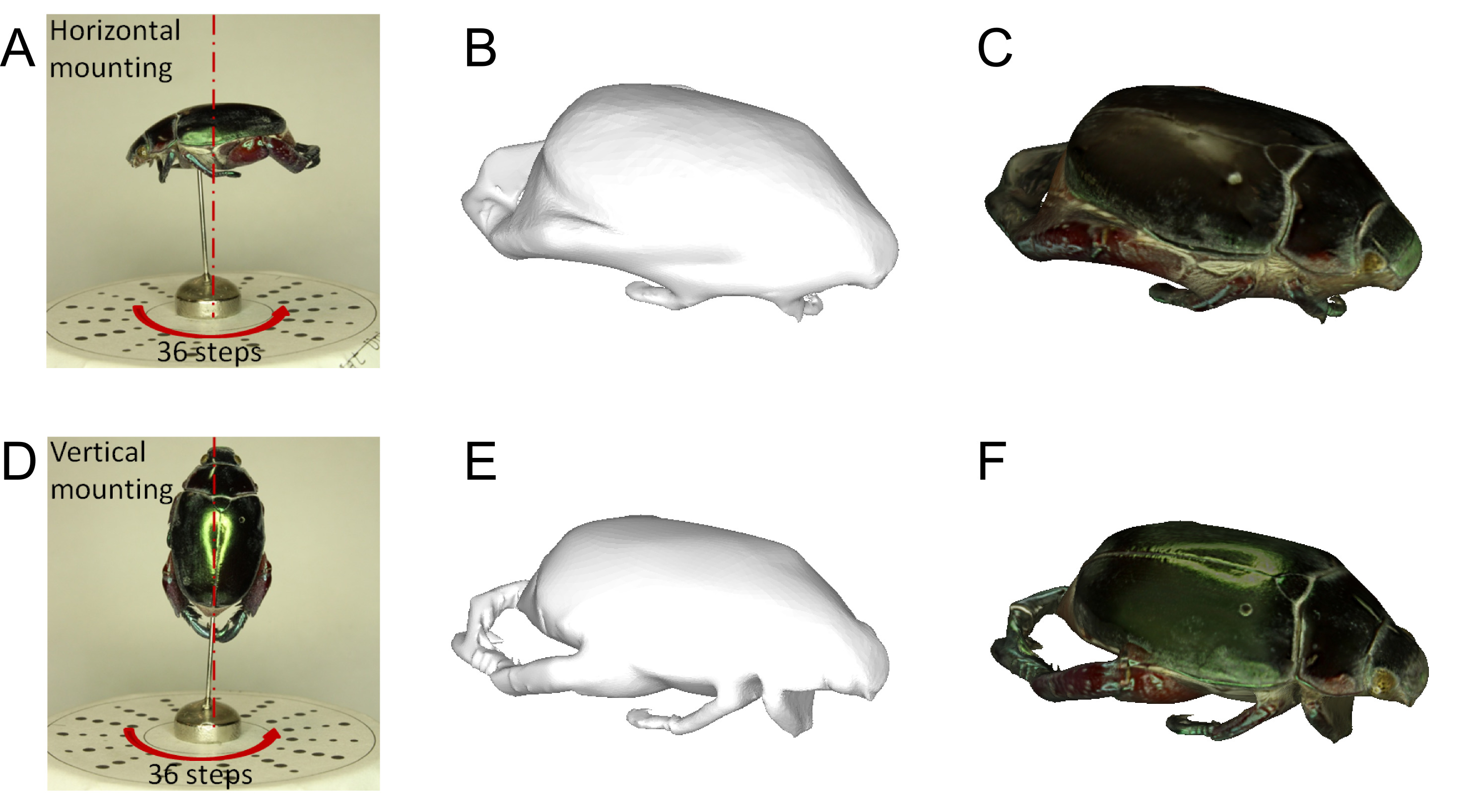}} 
\end{center}
\caption{\label{fig:mount_no_tilt} \textbf{The impact of mounting orientation on 
reconstruction quality.} Traditional horizontal mounting (A-C) produces inferior 
results to vertical mounting (D-F) for this specimen.}
\end{figure}

\begin{figure}[!ht]
\begin{center}
\textbf{\includegraphics[width=12.35cm]{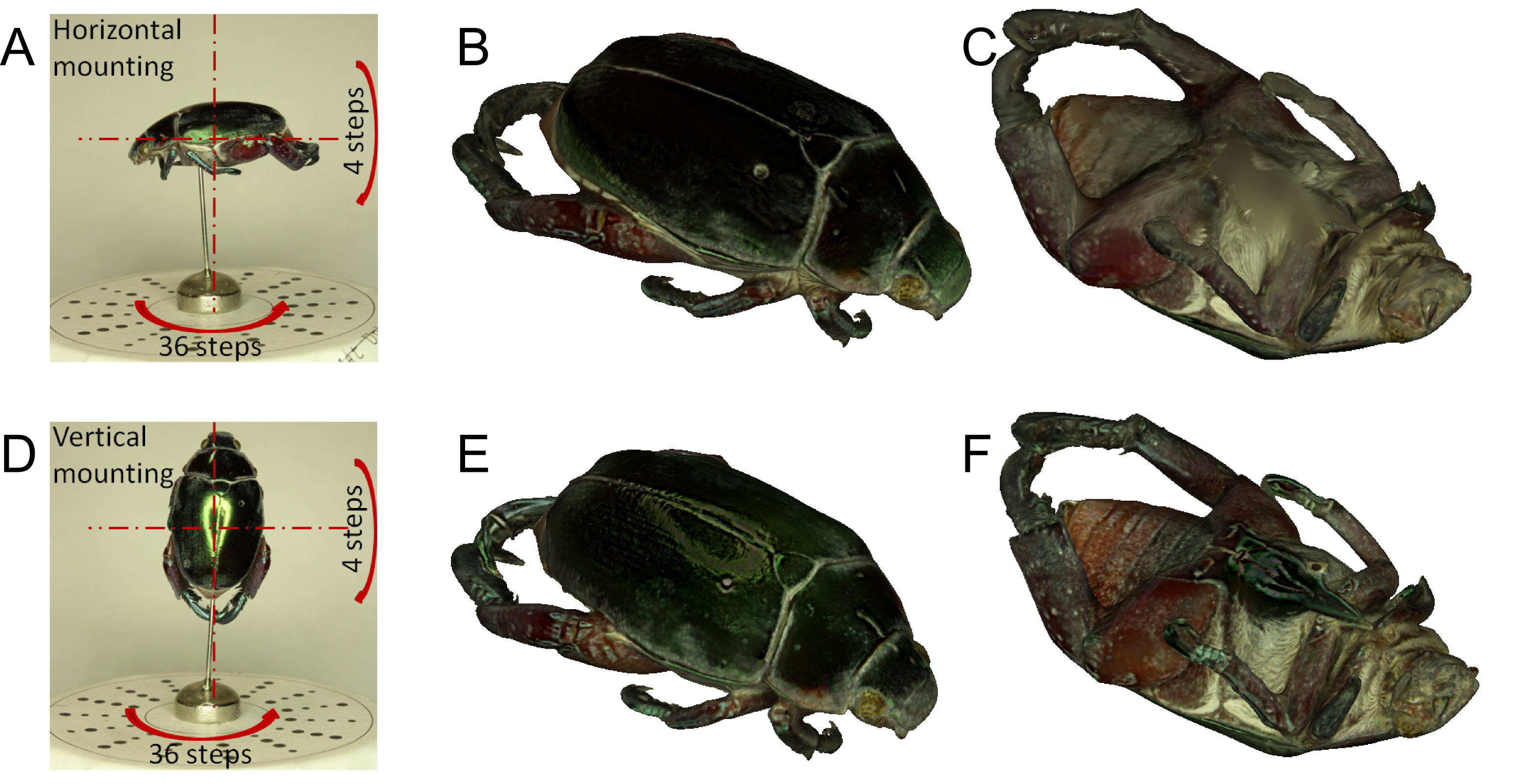}} 
\end{center}
\caption{\label{fig:mount_and_tilt} \textbf{Impacts of mounting orientation and
tilt on reconstruction quality.} While additional images at 
tilting angles of $10^0$, $20^0$, $30^0$ and $40^0$ improve reconstruction
quality in both horizontal and vertical mounting (in comparison with Figure~\ref{fig:mount_no_tilt}),
vertical mounting leads to sharper model with more vivid colours and textures. }
\end{figure}

By convention, insect specimens are often mounted horizontally. 
However this mounting orientation may not be ideal for 3D reconstruction.
To investigate the effect of mounting orientation on reconstruction
quality, we acquired images of a specimen mounted horizontally, then
vertically (Figure~\ref{fig:mount_no_tilt}). For the structure of
that particular specimen, vertical mounting gave markedly better reconstruction
of both geometry and colour, avoiding occlusions and capturing texture
in more detail. Increasing the number and variety of poses by acquiring
images at different tilt angles improved the reconstructions of both
vertically and horizontally mounted insects (Figure~\ref{fig:mount_and_tilt}).
Even in this case, vertical mounting afforded more detail in geometry and colour.
We therefore note that the best mounting orientation is specimen dependent:
visual hull reconstruction of geometry improves the more surface normals
are captured in silhouette, while colour and texture improve the more surface
normals are captured parallel to the camera viewing axis.

\begin{figure}[!ht]
\begin{center}
\textbf{\includegraphics[width=12.35cm]{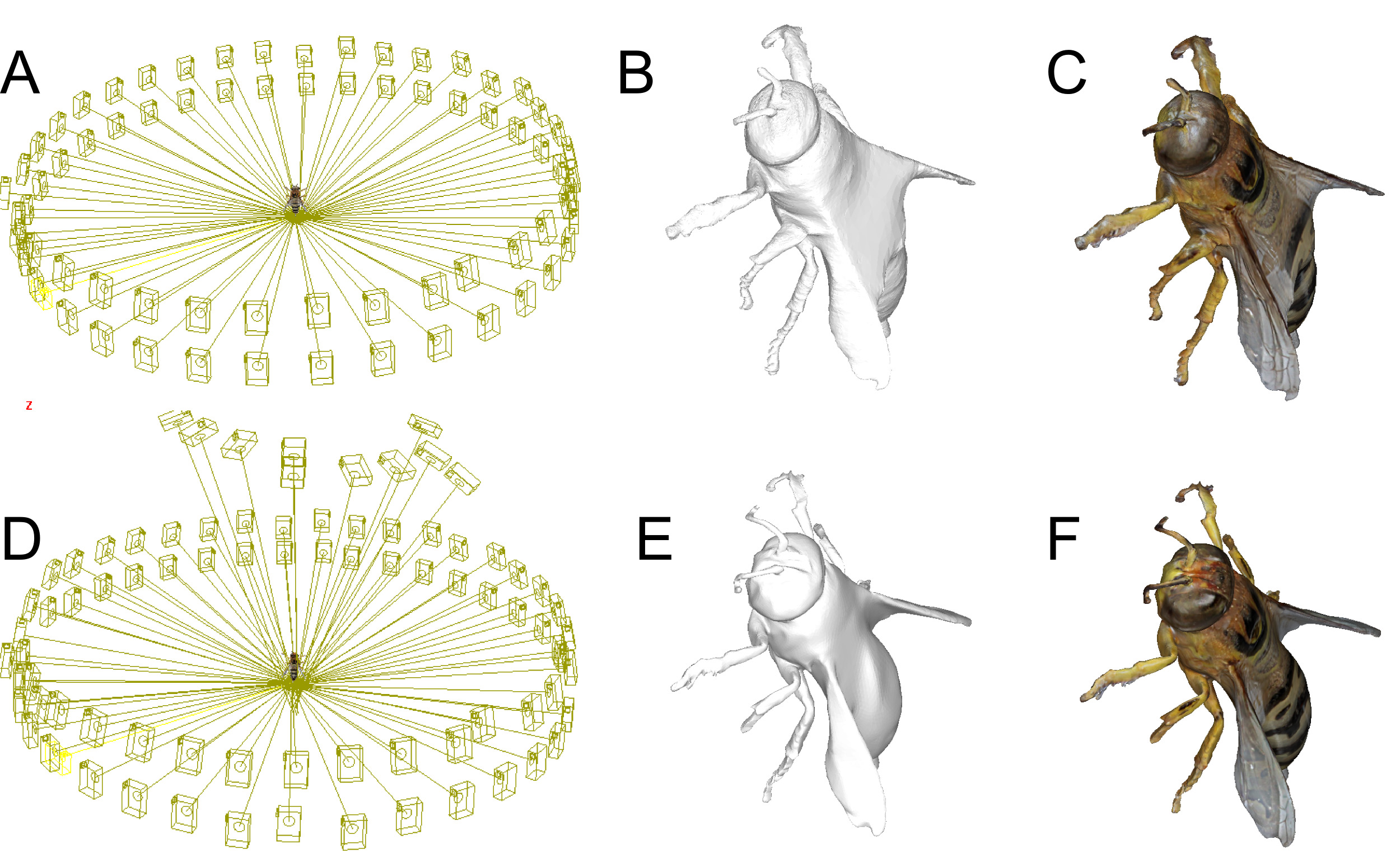} } 
\end{center}
\caption{\label{fig:insect-with-wings} \textbf{Additional camera poses can improve
wing reconstruction.} A) A typical set of camera poses cannot resolve the 
occlusion created by the wings of this insect, leading to inaccurate reconstruction 
between its wings (B-C). D) Additional images taken from camera poses looking 
along the insect body and wing surfaces dramatically improves reconstruction 
accuracy (E-F).}
\end{figure}

Further surface geometry issues arise as the structures of
specimens become more complex. Wings, for example, can be especially
challenging as shown in Figure~\ref{fig:insect-with-wings}(A-C)
where self-occlusion causes poor reconstruction of the wings. Fortunately,
additional informative views can be obtained to alleviate this problem
(Figure~\ref{fig:insect-with-wings}D-F).
Ideally, some of these additional views will be captured tangentially
to the wing surface to ensure the reconstructed wings have the correct
thickness.

\begin{figure}[!ht]
\begin{center}
\textbf{\includegraphics[width=12.35cm]{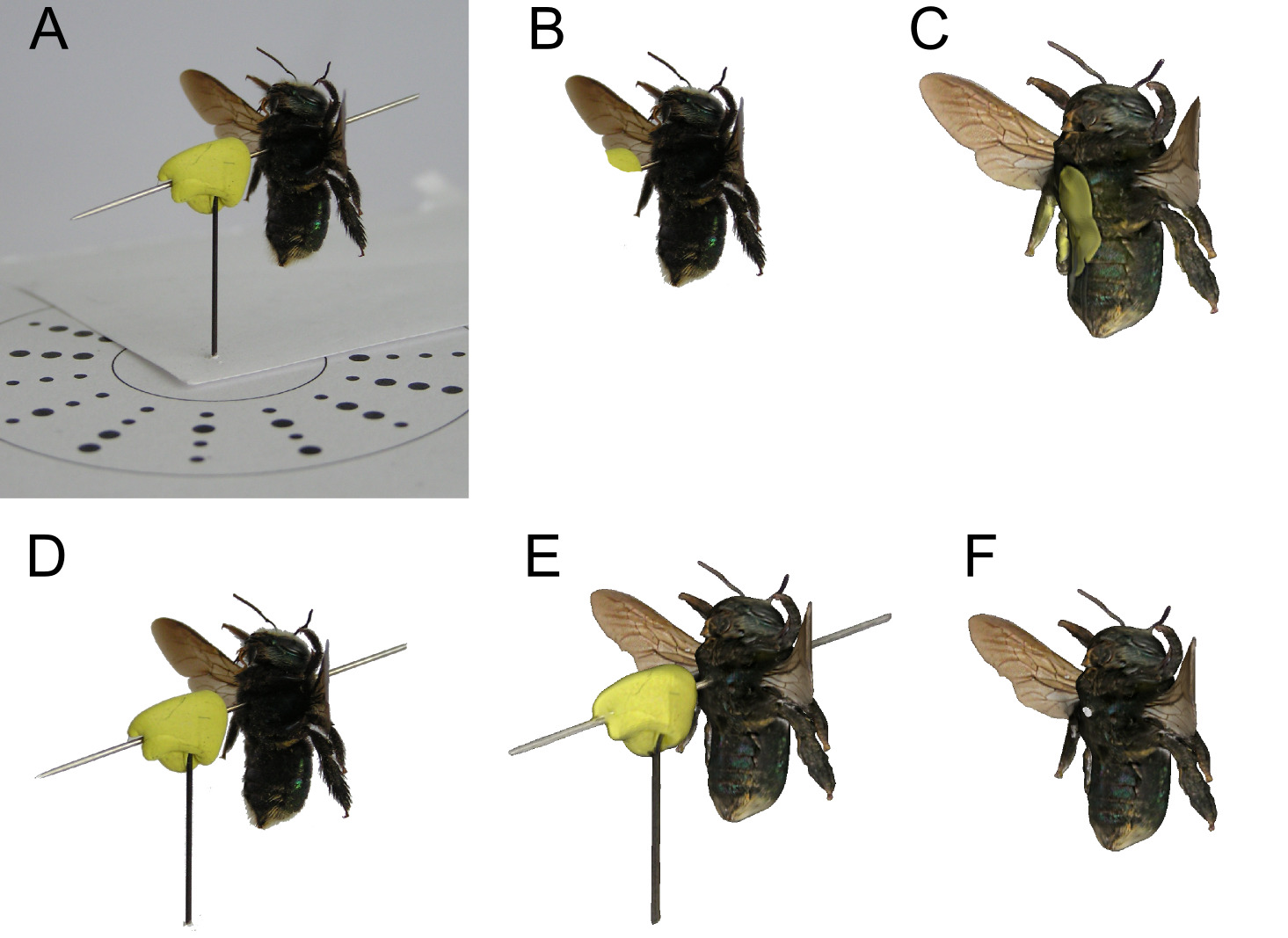}} 
\end{center}
\caption{\label{fig:non_removable_pin} \textbf{Two methods to deal with an insect whose
pin cannot be removed.} A) The raw image shows the pinned specimen attached to a second 
vertical pin so the long-axis of the insect is vertical. B) An image of 
the specimen after all other parts of the image are masked to some extent. C) Ventral view 
of the 3D reconstruction from masked images shows a splotch of contaminated
texture colour. D) An image of the specimen and pins retained. 
E) 3D reconstruction of insect and pins. F) Ventral view of E with pins edited
out of the 3D model.}
\end{figure}

We explored ways to achieve an informative mounting orientation
even when the specimen cannot be re-pinned (e.g., when the specimen
is too precious to handle, or the pin too firmly embedded to remove
without certain damage). Previously, we mentioned that 
vertical orientation provides better quality than the horizontal orientation.
However, repinning the specimen to have a vertical orientation causes damage, 
while keeping the horizontal orientation produces a lower-quality 3D model.
To avoid this trade-off,  the normally-pinned insect can be attached 
to a second pin (in this case using yellow Blu-Tack) so that the
specimen is rotated on its long axis (Figure~\ref{fig:non_removable_pin}A).
Then, the pins and the Blu-Tack need to be removed digitally 
to produce a clean final 3D model of the specimen.
There are two methods to do this.
The first method involves editing the Blu-Tack and mounting pins out
of the set of 2D images (Figure~\ref{fig:non_removable_pin}B) during background removal
 prior to reconstruction. However, this method does not work well with 
image views where the pins and Blu-Tack occlude
parts of the insect and the resulting reconstruction shows contaminated
texture colour (Figure~\ref{fig:non_removable_pin}C).
The second method is to keep the pins and Blu-Tack with the specimen
during 3D reconstruction (Figure~\ref{fig:non_removable_pin}D and E)  
\emph{then} remove them from the 3D model
using a mesh editor. Overall, this second strategy produces the better
result (Figure~\ref{fig:non_removable_pin}F).

In this paper, we have shown that high
resolution, natural-colour 3D digitization system for insects and other
small specimens can be implemented using readily available components with hardware
and software cost under AUD8000. As well as being cost effective,
the system produces digital 3D models that are fairly efficient in terms
of the ratio of information to data. The file size of the 3D granary
weevil model shown in Figure~\ref{fig:aperture_vs_multifocus}H
is around 10 megabytes. It was reconstructed from 18 megapixel
2D JPEG images (2 -- 4 megabytes/image) taken at 144
different angles and 31 different distances creating 10 -- 17 gigabytes
of 2D image data in all for a single specimen. By stacking 
each set of 31 multi-focus images into a single in-focus one, the image data is 
reduced approximately 20 times. By transforming this 2D data into a 3D model,
the system further achieves a 30:1 compression of data. This level of compression enables
useful information about the specimen to be exchanged via email, presented
in web pages and embedded in 3D PDF documents.

This work raises a number of research challenges
and opportunities for further improvement, including:
\begin{itemize}
\item Eliminating the need for the printed mat: 3DSOM\texttrademark{} requires
this mat to estimate the camera pose of individual images. We have
reached the lower size limit of what we can straightforwardly print
and attach to specimens. Furthermore, the range of poses is limited
to those in which the mat is viewable. There are reconstruction methods
that do not need this kind of pattern to estimate camera pose (e.g.,
\cite{Agisoft2013, Autodesk2013}), relying instead on feature matching and bundle
adjustment. However, the accuracy of these estimates depend strongly
on the geometry of the specimen and other objects captured in the
images.

\item Detailed features, such as hairs and surface roughness, demand higher
2D image and 3D model resolution and a concomitant increase in the
memory and computation needed to store and visualize the model. Our
strategy is to leverage the high resolution 2D image corresponding to
a particular pose of interest, reminiscent of the approach used in
\cite{Ortery2013}

\item Concave surfaces: current photo-consistency based methods to resolve
concavities can be challenged by the specular reflective properties of
many insects.

\item Transparent wings and membranes pose challenges for acquisition, reconstruction,
and for representation and rendering of the resulting 3D model.

\item View- and lighting-dependent appearance such as iridescence or 
sub-surface light scattering is also difficult to capture, represent and render.

\item 3D annotation standards, strategies and software are not yet as developed
as 2D approaches. The ability to augment 3D models with additional
information is important for taxonomy and other scientific ends, as
well as engaging a broader range of end users.

\end{itemize}

Despite these future challenges, we believe that the proof-of-concept prototype
presented in this paper demonstrates that natural-colour 3D model digitization
is feasible and affordable enough for insect collections to implement
and apply right now. 

An initial investigation of the usefulness of 3D insect models,
as described in the \emph{Supplementary Information} section,
showed that the quality of 3D insect models were good enough to
provide sufficient information for species identification, and allow for easier specimen 
examination than the actual specimen being viewed under a microscope.

The specific usage scenarios for wider communities such as quarantine officer or educator.
A quarantine officer can use 3D models of invasive insects while on duty 
to improve the speed and the accuracy of identification process. 
The challenges and possible solutions by using 3D models in quarantine control 
have been discussed in \cite{Nguyen2013c}. For educators, 3D models of insects can be used
as rich education materials, allowing students to interact with  insects 
without the need to  access to fragile specimens.

\section*{Supplementary Information}

\subsection*{Traditional insect mounting}

\begin{figure}[!ht]
\begin{center}
\includegraphics[width=3in]{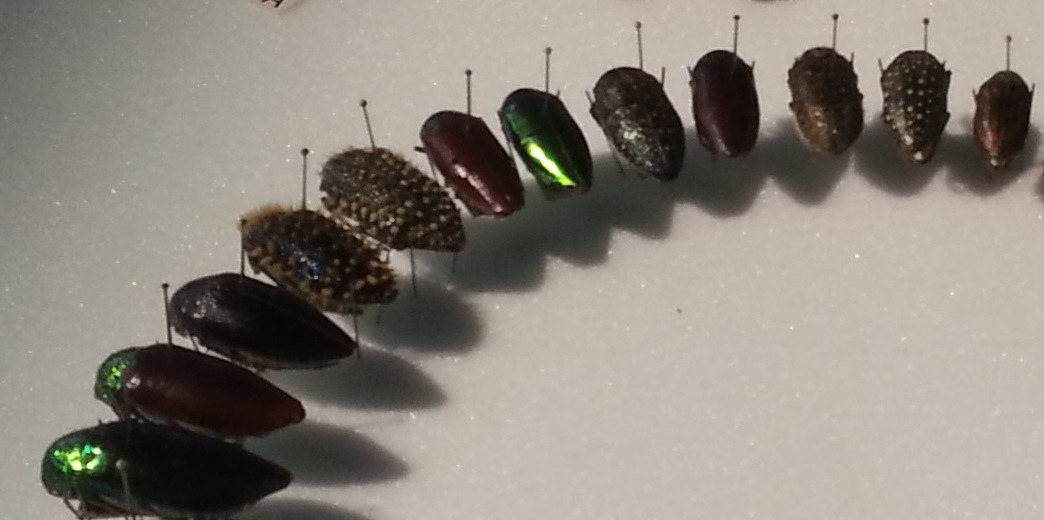} 
\end{center}
\caption{\label{fig:horiz_mounting} \textbf{Traditional insect mounting.} A metal
pin pierces the body of an insect from the back providing stronghold to the specimen. 
However this mounting can be inconvenient for 3D scanning due to the speciment's horizontal body axis.}
\end{figure}

Figure \ref{fig:horiz_mounting} shows insects mounted horizontally 
with a pin going through the body from the back. This mounting technique gives 
a strong hold on the insect body and facilitate specimen handling  
but provides limited access for 3D scanning.

\subsection*{Image processing and 3D reconstruction software}

Figure \ref{fig:image_pipeline} shows the overview of image processing and 
3D reconstruction pipeline. Software used in this process are: 
\begin{itemize}

\item Helicon Focus \cite{HeliconSoft2013} for multi-focus image stacking
to extend depth of field. There are alternative open-source software
such as CombineZP \cite{Hadley2010} and Huggins \& Enfuse \cite{D'Angelo2013}.
However we found Helicon easier to use and able to exploit multi-core processing. 

\item 3DSOM\texttrademark{} software \cite{AdamBaumbergCreativeDimensionSoftwareLtd}
for 3D reconstruction from multiple view images based on visual hull
technique. The 3D reconstruction pipeline used by 3DSOM\texttrademark{} is described
in \cite{Baumberg2005}. There are other 3D reconstruction software including commercial software 
 Agisoft Photo Scan \cite{Agisoft2013}; and open-source softwares Bundler \cite{Snavely2010} and Patch-based
Multi-view Stereo \cite{Furukawa2010a,Snavely2010},  and visual hull mesh
software \cite{Forbes2009}.
\end{itemize}
These two off-the-shelf software packages, together, cost around AUD1700 based on 2013 prices.

\subsection*{Image acquisition equipment}

\begin{figure}[!ht]
\begin{center}
\textbf{\includegraphics[width=6in]{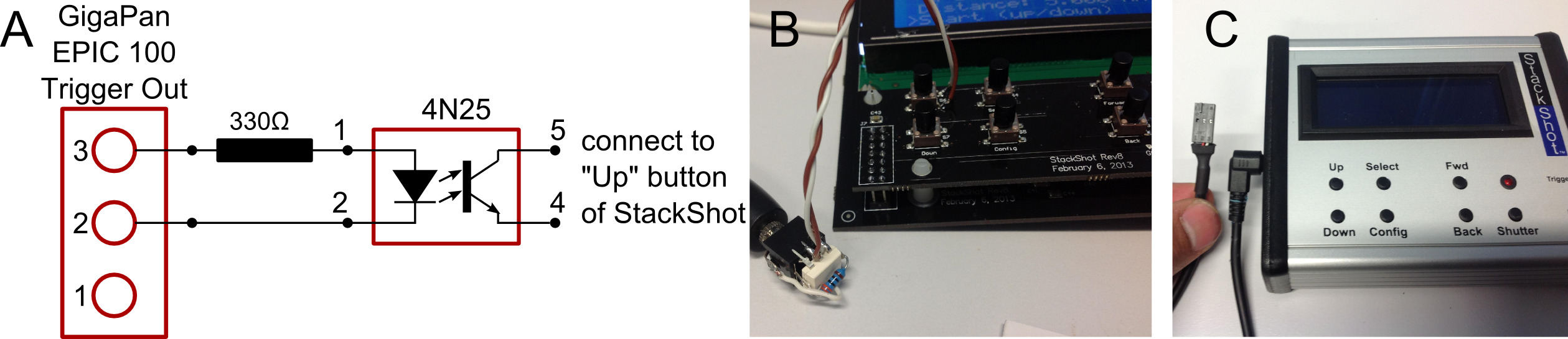} } 
\end{center}
\caption{\label{fig:StackShot} \textbf{Modification of StackShot\texttrademark{} macro-rail's control
box to accept external trigger.}
A) Circuit diagram to provide interface between trigger output
of GigaPan\texttrademark{} robot and StackShot\texttrademark{} control box. 
The resistor and the opto-coupler
convert the trigger signal to a ``press button'' action. B) the resistor and 
opto-coupler are soldered to a connector which can
be mounted to the case of the control box. C) Finished 
the control box with the extra input connector on the left. Note
that the connector pins of GigaPan\texttrademark{} trigger cable have to be swapped
such that red cable goes to position 2 and white cable to position
3. }
\end{figure}

We use the following off-the-shelf components: 

\begin{itemize}
\item A GigaPan\texttrademark{} Panorama Robot EPIC 100 \cite{GigaPanSystems2013}.
Normally this device is used to control the tilt and pan of a camera
to capture panorama images. In this project, we turn it side-ways and
use it as a two-axis turntable to control the tilt and pan of the
specimen to be imaged. The GigaPan\texttrademark{} Panorama Robot connects to a camera
via a cable and triggers a camera to capture one or more images
per rotation angle. 

\item A Canon\texttrademark{} 600D camera or better.
It provides low-noise, high
resolution images (18MP). It is highly customizable with several
ports for external trigger input, flash trigger output, and a USB port for
remote tethering. One problem with many DSLR cameras is that they 
rely on a mechanical shutter which can cause mechanical vibration and
image blur. The image blur due to mechanical
shutter vibration can be alleviated by using a camera flash triggered between shutter movement. 

\item A Canon\texttrademark{} EF 100mm macro lens for normal-mode 
and Canon\texttrademark{} MP-65mm for macro
mode. The EF-100mm lens provides magnification of $1\times$ or smaller, while
the MP-E 65mm lens provides magnification from $1\times$ to $5\times$. These macro
lenses also have much less optical distortion than some other
lenses. A low-cost solution to make normal lens achieve higher
magnification is to add a macro extension
tube between the camera and the lens. However this often increases optical distortions 
\cite{McHugh2013}. The depth of field of MP-65mm lens as function of 
magnification and aperture can be obtained from its user manual.

\item A Viltrox\texttrademark{} JY-670 macro ring flash \cite{JVCTechnology2013}. This
flash provides illumination and reduces 
exposure time. Short exposure time is critical for macro photography
to reduce image blurring due to vibration from the camera's mechanical
shutter. Additionally, a Tronix SpeedFire flash power supply \cite{InnovatronixInc2013}
is used for fast flash charging. 

\item A StackShot\texttrademark{} macro-rail \cite{CognisysInc2013}. This
device enables us to capture high-magnification partially-focused images at predetermined depth intervals
with high position accuracy. This macro-rail can run automatically to capture
images along a single direction. The process of movement and image
acquisition using macro-rail starts with a press of ``Up'' button
on the rail's control panel. To synchronize StackShot\texttrademark{}
macro-rail and GigaPan\texttrademark{} robot, we built a circuit interface, shown
in Figure \ref{fig:StackShot}, to convert the robot's trigger signal
to a press-button effect. The step size of the rail is set to be approximately 70\% 
of the depth of field of the macro lens to allow for adjacent overlapping required for stacking multi-focus images.

\item Two 1mW laser pointers \cite{Dick-Smith2013} for specimen alignment. 
The laser beams are used to locate the specimen at the center of rotation of the 
two-axis turntable so that the specimen stays at the same position while 
being rotated and imaged.

\item Aluminium frames \cite{Bunnings2013a} to hold the turntable, laser
pointers, camera, and macro-rails in position. 

\item Button magnets \cite{FrenergyMagnets2013}, plastic spacers \cite{Bunnings2013},
epoxy glue and pins to mount insect specimens. 
\end{itemize}

The system does not need any special software for acquiring images.
The EOS Utility software accompanying the camera is needed to remote-tether
the camera and transfer images to a computer during image acquisition.
Based on 2013 prices, we estimate the total hardware to cost around AUD3200
to AUD5600.

Flow charts of how the image acquisition system works in normal and
macro-modes are shown in Figure \ref{fig:flowcharts}. These charts
are generalized so that they are hardware independent. ``Pan rotation''
refers to rotation around the vertical axis of the specimen 
and ``tilt rotation'' refers to rotation around horizontal axis 
(perpendicular to both the vertical axis and the camera-specimen axis). 

\subsection*{Estimation of processing time}

The reconstruction process described in this paper is a proof of concept and 
not yet optimized for speed and for large-scale digitization. A time estimation for 
each stage of the 3D reconstruction procedure for the insects shown in Figure  \ref{fig:bugs}
is provided in Table \ref{tab:time_estimation}.
The longhorn beetle, Christmas beetle and Amycterine ground weevil take relatively similar 
amount of time to obtain a 3D model. The sand wasp has increased reconstruction time 
due to the extra time required to manually correct the errors of background removal 
around the wings,  and to refine camera pose during reconstruction 
(hairs significantly increase the time for pose refinement). 
The granary weevil, due to its small size, takes much more time to mount using a  microscope, 
acquire multi-focus images and perform 3D reconstruction. 
It is often the case that the reconstruction stage is repeated several times to find and fix errors 
in background removal, or to iterate the refinement process until the result is acceptable.
Therefore, the times provided in Table \ref{tab:time_estimation} should only be considered 
as indicative.

\subsection*{Informal feedback on using a 3D insect model for species identification}

One possible use case for 3D insect models is species identification.
In drafting this paper, we needed
to fully identify the Christmas beetle shown in Figure~\ref{fig:bugs}.
We asked CSIRO entomologist Mr Tom Weir to try to do
so from the photograph, the 3D model, and finally the actual specimen. 
In addition to identifying the specimen as a male \emph{Anoplognathus viriditarsus}
from the 3D model alone, Mr Weir provided feedback which we summarize as:

\begin{itemize}
\item The single top-view photograph does not capture the key features needed
to identify the species, just enough information to determine its
genus. The features are located at different parts of the insects
such as abdomen, head, mouth, claws, rear end, etc. Multiple images
captured at particular angles are required to show all these features.

\item The 3D model provides useful extra information and allows views from
any angle. There are still missing features such as the mouth area
and hairy surface on the head/nose due to resolution of geometries
and texture colours. However the available features are roughly enough
to identify its species. Examination and species identification would
be facilitated if the texture information was provided more clearly,
or high-resolution images of the features were attached to the model.

\item The actual specimen obviously provides all the information needed
but needs examination under a microscope for some features such as
the mouth area and hair surface on the head. However, out-of-focus 
effect and other physical restrictions makes the use of microscope to view 
the actual specimen more cumbersome 
than to view the 3D model of specimen on a mobile device (such as iPad).

\item Identification involves matching the given specimen to its corresponding
specimen in the Australian National Insect Collection \emph{and} comparing
it to closely related species.

\item For identification purposes, the natural-colour 3D model is much more
helpful than the Micro CT 3D model. The better geometry accuracy of
the Micro CT model does not assist in this instance.

\item Higher resolution of the natural-colour 3D model is desirable to provide details
of key identification features. This is particularly important for
features, such as hairs, that never get included in the 3D models.
One solution would be to attach high resolution photographs to locations
of the key features. 

\item To make the 3D model useful for species identification, it is important
to know in advance all key features and the identification procedure
and level. Ideally, this information would be arranged in a way that
directly supports fast, correct identification. 3D visualization and
annotation could make the identification procedure more obvious that
the instructions currently provided in text and 2D illustrations.
\end{itemize}

\section*{Acknowledgements}

We thank Ms Eleanor Gates-Stuart (CSIRO Computational Informatics), 
Dr Rolf Oberprieler, Ms Jennings Debbie,
Dr Beth Mantle, Ms Nicole Fisher, Ms Cate Lemann, Ms Anne Hastings, Mr Eric
Hines and Mr Tom Weir (CSIRO Ecosystem Sciences) for their kind assistance
during the course of the project. Special thanks goes to Dr Sherry Mayo 
(CSIRO Materials Science \& Engineering) for providing 
Micro CT model of the granary weevil for comparison. 
We also thanks Dr Vince Smith (Natural History Museum - London), 
Prof Pavel Stoev (National Museum of Natural History - Sofia) 
and another anonymous reviewer for their valuable feedback to our manuscript.
We gratefully acknowledge CSIRO's
Office of the Chief Executive Postdoctoral Scheme, Transformational
Biology initiative, and Business and Service Analytics Program for
funding this work.

\section*{Author Contributions}

Created, guided and informed the project: DRL JLS MA.
Conceived and designed the methods and the experiments: CVN. 
Performed the experiments and analyzed the data: CVN. 
Wrote the paper: CVN DRL JLS MA.

\bibliographystyle{plos2009}
\bibliography{Chuong_3D_Insects}

\section*{Figure Legends}

\begin{figure}[!ht]
\begin{center}
\textbf{\includegraphics[width=6in]{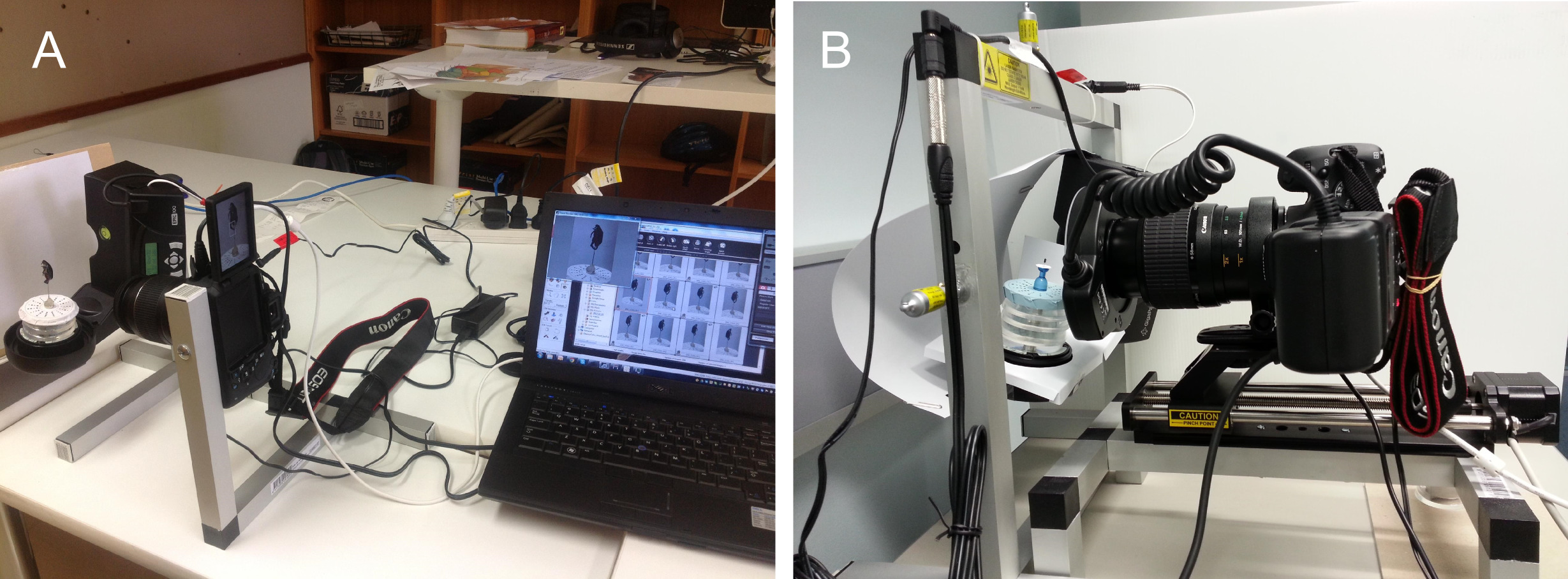} } 
\end{center}
\caption{\label{fig:setup} \textbf{Examples of actual set-up in normal-mode (A)
and macro-mode (B).} The GigaPan\texttrademark{} Panorama Robot is mounted sideways
to act as a two-axis turntable. A) camera is turned $90^0$
around optical axis so that it can capture images that better fit
the specimen and the mat target. 
Video of the system at work are available at \cite{Nguyen2014i}. }
\end{figure}

\section*{Tables}

\begin{table}[!ht]
\caption{
\bf{Time consumption estimation}}
\begin{tabular}{|p{3cm}|c|c|c|}
    \hline
    Insect & Mounting (min) & Acquisition (min) & Reconstruction (min) \\ \hline
    Longhorn beetle, Christmas beetle and ground weevil & 5 & 15 & 60 \\ \hline
    Sand Wasp & 5 & 15 & 120 \\ \hline
    Granary Weevil & 30 & 180 & 120 \\
    \hline
\end{tabular}
\begin{flushleft}
\end{flushleft}
\label{tab:time_estimation}
 \end{table}

\end{document}